\RequirePackage{fix-cm}

\documentclass{article}

\usepackage{arxiv}

\usepackage{graphicx}
\usepackage{mathptmx}      % use Times fonts if available on your TeX system
\usepackage{cite}
\usepackage[export]{adjustbox}
\usepackage{siunitx}
\usepackage{subcaption}
\usepackage{xcolor}
\usepackage{hyperref}
\usepackage{array}
\newcolumntype{P}[1]{>{\centering\arraybackslash}p{#1}}
\newcommand{\superscript}[1]{\ensuremath{^\textrm{#1}}}
\newcommand{\revised}[1]{\textcolor{black}{#1}}

\begin{document}
\title{A Learning-based Method for Online Adjustment of C-arm Cone-Beam CT Source Trajectories for Artifact Avoidance}
%\thanks{The auth}

%\titlerunning{Online Adjustment of C-arm Cone-Beam CT Source Trajectories}        % if too long for running head

\author{Mareike Thies\superscript{1,2,*}\\ \And Jan-Nico Z{\"a}ch\superscript{3,*}\\ \And Cong Gao\superscript{1}\\ \And Russell Taylor\superscript{1}\\ \And Nassir Navab\superscript{1}\\ \And Andreas Maier\superscript{2}\\ \And Mathias Unberath\superscript{1}\\
}

\footnotetext[1]{Laboratory for Computational Sensing + Robotics, Johns Hopkins University, Baltimore, MD}
\footnotetext[2]{Pattern Recognition Lab, Friedrich-Alexander-Universit{\"a}t Erlangen-N{\"u}rnberg}
\footnotetext[3]{Computer Vision Laboratory, Eidgen{\"o}ssische Technische Hochschule Z{\"u}rich}

\renewcommand{\thefootnote}{\fnsymbol{footnote}}
\footnotetext[1]{Both authors contributed equally and are listed in alphabetical order.}

%\institute{\email{unberath@jhu.edu}\\
%        \superscript{1} Laboratory for Computational Sensing + Robotics, Johns Hopkins University, Baltimore, MD\\
%        \superscript{2} Pattern Recognition Lab, Friedrich-Alexander-Universit{\"a}t Erlangen-N{\"u}rnberg\\
%		\superscript{3} Computer Vision Laboratory, Eidgen{\"o}ssische Technische Hochschule Z{\"u}rich\\
%		\superscript{*} Both authors contributed equally and are listed in alphabetical order.
%}

%\authorrunning{Thies and Z{ae}ch et al.} % if too long for running head

%\date{}

\maketitle

\begin{abstract}

\textit{Purpose:}
During spinal fusion surgery, screws are placed close to critical nerves suggesting the need for highly accurate screw placement. Verifying screw placement on high-quality tomographic imaging is essential. C-arm Cone-beam CT (CBCT) provides intraoperative 3D tomographic imaging which would allow for immediate \revised{verification} and, if needed, revision. However, the reconstruction quality attainable with commercial CBCT devices is insufficient, predominantly due to severe metal artifacts in the presence of pedicle screws. These artifacts arise from a mismatch between the true physics of image formation and an idealized model thereof assumed during reconstruction. Prospectively acquiring views onto anatomy that are least affected by this mismatch can, therefore, improve reconstruction quality.

\noindent \textit{Methods:}
We propose to adjust the C-arm CBCT source trajectory during the scan to optimize reconstruction quality with respect to a certain task, i.e. verification of screw placement.  Adjustments are performed on-the-fly using a convolutional neural network that regresses a quality index over all possible next views given the current x-ray image. Adjusting the CBCT trajectory to acquire the recommended views results in non-circular source orbits that avoid poor images, and thus, data inconsistencies.

\noindent \textit{Results:} 
We demonstrate that convolutional neural networks trained on realistically simulated data are capable of predicting quality metrics that enable scene-specific adjustments of the CBCT source trajectory. Using both realistically simulated data as well as real CBCT acquisitions of a semianthropomorphic phantom, we show that tomographic reconstructions of the resulting scene-specific CBCT acquisitions exhibit improved image quality particularly in terms of metal artifacts. 

\noindent \textit{Conclusion:}
The proposed method is a step towards online patient-specific C-arm CBCT source trajectories that enable high-quality tomographic imaging in the operating room. Since the optimization objective is implicitly encoded in a neural network trained on large amounts of well-annotated projection images, the proposed approach overcomes the need for 3D information at run-time.

\end{abstract}

\section{Introduction}
\label{sec:intro}
The number of patients undergoing spinal fusion surgery in the US has been increasing rapidly over the last years. From 204,000 cases in 1998, the number of interventions has grown to 457,000 cases in 2011. This growth also resulted in a steep increase in hospitalization charges of more than 750\% as each intervention is relatively expensive with an average hospital bill of more than \$34,000 \cite{andersson2014,deyo2004}. During spinal fusion surgery, two or more vertebrae are fused. Usually, this is achieved by implanting screws into the affected vertebrae in a transpedicular approach. These screws are then interlocked with metal rods to inhibit movement and induce the formation of new bony structure. This stabilizes the spine at the given location, which can reduce chronic back pain when non-operative treatment failed \cite{szpalski2010}. 
Despite the high number of interventions, spinal fusion surgery remains a high-risk operation. In 2010, 6.8\% of the patients undergoing spinal fusion interventions in the US were rehospitalized within the first 30 days after surgery \cite{weiss2013}. Misplacement of pedicle screws has been reported in up to 55\% of the cases using traditional free-hand technique for screw placement \cite{manbachi2016}. While this number can be decisively reduced when the intervention is performed under fluoroscopy-guidance \cite{fu2008}, even in navigated approaches the incidence of misplaced screws remains high \cite{gelalis2012accuracy}. Screw misplacement intrinsically carries the danger of cortical breach which can result in nerve damage and severe neurological impairment of the patient \cite{thomsen1997, fritzell2003} and has been found in up to 8\% of the cases \cite{manbachi2016}.
Consequently, there is a need for an imaging modality capable of precisely capturing the anatomy in direct proximity of metal implants to assess the adequacy of implant placement \emph{during} the intervention, allowing for immediate revision in case of misplacement. 
Tomographic reconstructions available through C-arm cone-beam CT (CBCT) have the potential to provide such information intraoperatively. CBCT has been deployed for this purpose \cite{sembrano2012, garber2012}. However, even with one of the most recent CBCT devices and compared to conventional postoperative CT, 23\% of the cases of cortical breach were missed on intraoperarive CBCT images primarily due to much stronger metal artifacts around the screw \cite{cordemans2017}. Improving the quality of intraoperative CBCT reconstructions for the task of pedicle screw placement in spinal fusion surgery, consequently, has a great potential to identify cortical breach during the operation, allowing for immediate revision, and therefore, reduction of both neurological complications and need for revision surgery.

\revised{\subsection{Background}}
Artifacts in CT reconstructions are a consequence of discrepancies between the assumed mathematical forward model of x-ray image formation and the real physics of data acquisition \cite{schulze2011}. The process of reconstruction aims at finding a volumetric representation that optimally explains all measured x-ray images which is the inverse problem to the image formation process. Usually, this is performed by backprojecting the images into the volume thereby inverting the idealized mathematical assumption of the forward process \cite{feldkamp1984}. This is a well-posed problem as long as the mentioned discrepancies remain small. High discrepancies, however, render this an ill-posed inverse problem resulting in artifacts in the reconstructed volume. Other than noise, one of the most prominent discrepancies is beam hardening. It is characterized by a shift between incident and recorded energetic spectrum of the photons due to energy-dependent attenuation in dense objects, e.g. in our case titanium screws. This affects measurements on the detector and results in the overestimation of certain pixels during backprojection \cite{schulze2010}.  
Existing approaches to artifact reduction in CBCT reconstructions usually rely on postprocessing of the acquired data in projection domain \cite{zhang2007, meilinger2011, liao2019} or artifact suppression in volume domain \cite{kondo2010}. These methods work on the corrupted data and oftentimes carry the risk of introducing new sub-optimal image content which compromises the quality of the reconstruction in a different way. Therefore, we propose to begin artifact reduction one step earlier by adjusting the CBCT protocol to the scene and directly acquire better data. \revised{Specifically, our approach automatically predicts adjustments to the C-arm trajectory in real-time to actively exploit views onto the anatomy which are most consistent with the assumptions made in the tomographic reconstruction process.} Selection of good views is performed by a convolutional neural network that regresses a view-dependent quality index from the current projection of an ongoing scan. We hypothesize that this finally leads to artifact reduction and improved quality of the reconstruction. Ultimately, such approach could allow for intraoperative tomographic imaging with clinically acceptable quality in applications such as spinal fusion surgery.

\revised{\subsection{Related work}}
Conceptually related ideas to ours have been proposed for real-time user guidance in free-hand ultrasound probe motion. In \cite{milletari2019} the ultrasound image of each time point is interpreted by a deep reinforcement learning agent which predicts an incremental update on the probe motion. Similarly, incremental and real-time user feedback can be provided in the case of SPECT imaging with mobile freehand detectors based on the numerical condition of the system matrix corresponding to the reconstruction problem \cite{vogel2013}. However, analyzing the entire system matrix is not feasible for CT due to its memory footprint. Instead, different approaches have been proposed to select the most valuable next projection for CT: The method in \cite{zheng2011} favors views that sample rays tangential to edges of the 3D object to maximize the edge information in the reconstruction, obviously relying on precise knowledge of the object at optimization time. In \cite{dabravolski2014} angular steps are selected such that with each additional projection the set of solutions that are consistent with the set of already measured projections is minimized. This, however, has only been applied to the 2D reconstruction case and is computationally very expensive as it needs several preliminary reconstructions per step. Recently, finding an optimal sinusoidal trajectory which avoids metal parts of the imaged object while still ensuring a high coverage in Radon space for its direct vicinity has been studied in \cite{gang2020}. While all these approaches do not directly consider the x-ray imaging physics, \cite{stayman2013} proposes an index to analyze the quality of different projections based on the local point spread function and noise power spectrum of the imaging device. Similarly, \cite{wu2020} calculates a quality map of possible views from the expected amount of spectral shift due to beam hardening depending on different path lengths of the photons through metal objects. Both methods were successfully applied to CBCT trajectory optimization. Yet, all previous approaches calculate optimal parameters in a (semi-)offline manner and rely on knowledge about the 3D object at optimization time, which is usually provided by a preoperative scan. This requirement is problematic since, during interventions, the anatomy is altered in an unpredictable way, e.g. by screw insertion.

\revised{\subsection{Contribution}
In this work, we expand on our MICCAI 2019 submission that introduced machine learning-based algorithms to predict on-the-fly adjustments for task-based C-arm trajectories \cite{zaech2019}. 
Our contributions are two-fold. First, in carefully controlled experiments on synthetic data we characterize the algorithm’s behavior and robustness 1) in the presence of varied noise levels, and 2) with varying initial poses of the C-arm gantry with respect to anatomy. Second, we substantially expand our experiments on real data acquired from a semianthropomorphic phantom. To this end, we acquire 17 CBCT short-scans at different swivel and tilt angles of the gantry and align all projection images to a common 3D object space via image-based registration. This produces a set of calibrated x-ray images that allows for validating the proposed C-arm servoing algorithm in a retrospective manner. 
This strategy enables feasibility studies on real x-ray data but avoids the need for 1) a fully robotized and freely steerable C-arm device, and 2) flexible and robust online calibration methods that accurately estimate the C-arm imaging geometry without prior knowledge on the 3D scene. Exploring solutions to these challenges is important and will be the subject of our future work.}

%This work extends our previous paper on optimizing CBCT scan trajectories \cite{zaech2019} in the following aspects:
%\begin{enumerate}
%    \item Extended literature review about comparable methods 
%    \item Evaluation of the algorithm's robustness and behavior in the presence of noise and varying initialization poses of the C-arm
%    \item Data collection on a C-arm CT scanner that allows for calibrated reconstruction from non-circular trajectories
%    \item Extension of the algorithm such that optimal non-circular trajectories are composed from the collected data using the network predictions which allows for evaluation of the envisioned pipeline on real data
%\end{enumerate}

\section{Methods}

\subsection{Online Trajectory Adjustment Pipeline}
\label{sec:pipeline}
\begin{figure}
    \centering
    \includegraphics[width=\textwidth]{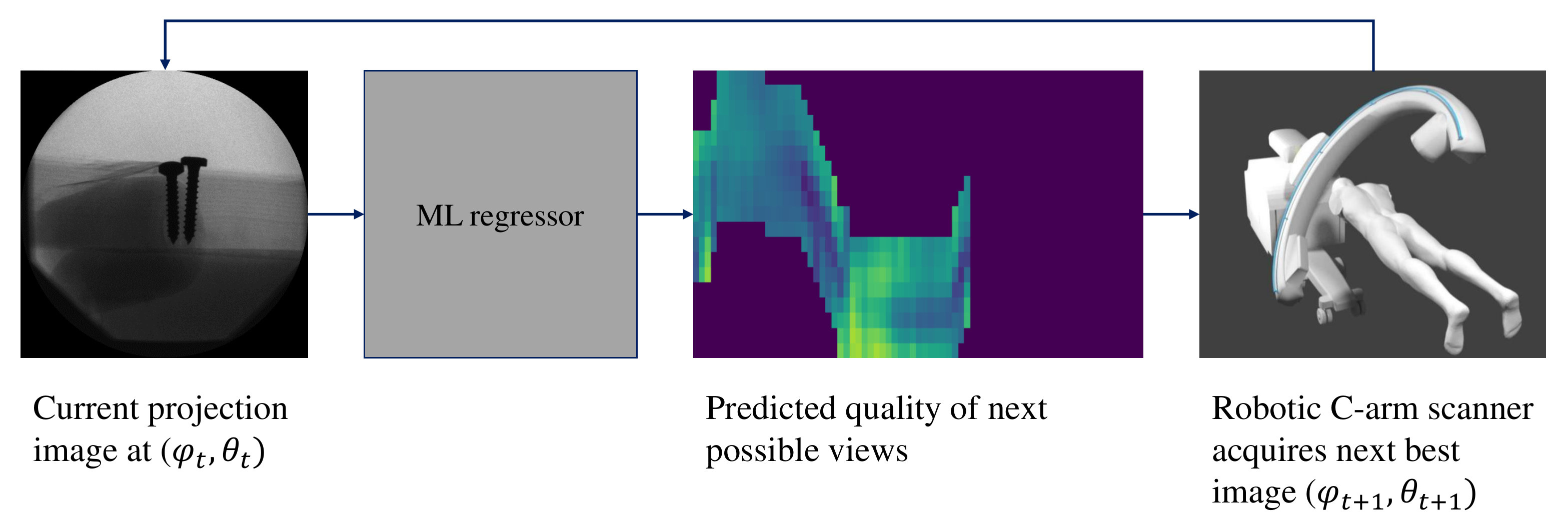}
    \caption{\revised{High-level overview of the envisioned pipeline for online trajectory adjustment. }}
    \label{fig:pipeline}
\end{figure}

Trajectory optimization is a problem with many degrees of freedom because recent scanners can realize very different motion patterns. Following the ideas in \cite{stayman2013}, we choose to parameterize the problem in terms of an in-plane angle $\varphi$ and an out-of-plane angle $\theta$. The in-plane angle is defined according to a traditional circular trajectory where source and detector move in one plane for the entire scan, whereas the out-of-plane angle is associated with tilting the C-arm relatively to this plane. Each trajectory consists of a set of pairs $(\varphi_t, \theta_t), \; t=0,..,T$ where $T$ is the total amount of projections images. 
The general pipeline we propose is illustrated in figure \ref{fig:pipeline}. An x-ray image is captured at a position $(\varphi_t, \theta_t)$ and processed by a VGG-type convolutional neural network, which regresses a detectability index (see section \ref{sec:index}) for the next possible projections. The projection with the highest predicted value is identified and the out-of-plane angle $\theta_{t+1}$ is updated accordingly while the in-plane angle $\varphi$ is always incremented by a fixed amount: $\varphi_{t + 1} = \varphi_t + \Delta\varphi$. The new target $(\varphi_{t+1}, \theta_{t+1})$ is sent to the robotic C-arm and its position adjusted accordingly to acquire the next projection. 

\subsection{Projection-dependent Detectability Index}
\label{sec:index}
To assess how single projections contribute to perceived reconstruction quality, we follow existing approaches based on the non-prewhitening matched filter observer model which allows to find a so-called detectability index as per equation \ref{eq:dsquared} \cite{gang2014}. 
\begin{equation}\label{eq:dsquared}
    d^2(\varphi, \theta) = \frac{[\int\int\int|\mathrm{MTF}(\varphi, \theta)|^2|\mathrm{W}_{task}|^2 \mathrm{d}f_x \mathrm{d}f_y \mathrm{d}f_z]^2}{\int\int\int\mathrm{NPS}(\varphi, \theta)|\mathrm{MTF}(\varphi, \theta)|^2|\mathrm{W}_{task}|^2 \mathrm{d}f_x \mathrm{d}f_y \mathrm{d}f_z}
\end{equation}
MTF is the local modulation transfer function, NPS is the local noise power spectrum and W$_{task}$ is a task-function describing the properties of the object to be imaged with highest quality in Fourier space. 
For the case of iterative penalized-likelihood reconstruction, it is possible to derive analytic expressions for both MTF and NPS \cite{gang2014}. These equations rely on forward projecting voxels into all views contained in a trajectory, comparing the projected value with the measured values and back projecting this information into the volume. Using these calculations for MTF and NPS, the final detectability index $d^2$ thus depends on the 3D structure of the imaged object as well as the set of images in a trajectory. This means that, if accurate 3D information is available, equation \ref{eq:dsquared} can be maximized with respect to $\varphi$ and $\theta$ to find an optimal trajectory. \revised{Note that the local MTF and NPS are very general measures which can be calculated for any imaged object. This work is centered around metal artifacts suppression as these are usually the most severe artifacts during interventions. In a different setting, the same index could potentially also be used for improving e.g. soft-tissue contrast.}

\subsection{Network for Detectability Prediction}
\label{sec:network}
During an intervention, the volume to be imaged is altered compared to preoperatively acquired information. Therefore, offline trajectory optimization approaches (e.g. the one outlined in Section~\ref{sec:index}) usually cannot succeed in these cases. As introduced in our previous work \cite{zaech2019}, we instead propose to regress the detectability index in equation \ref{eq:dsquared} on-the-fly during an ongoing scan using a convolutional neural network \revised{(CNN)} using only fluoroscopic images as input. In this approach, knowledge about the task is encoded in the weights of the machine learning model, thereby overcoming the need for explicit 3D information at CBCT acquisition time.
\revised{We rely on an architecture that is similar to the VGG architecture \cite{simonyan2014}, but adapted to perform regression instead of classification because we believe that a highly parameterized CNN is well suited to implicitly capture the underlying 3D structure in a learning-based manner.} From an input x-ray projection, the network is trained to predict the detectability of those projections with an increment of $+5^{\circ}$ in in-plane angle and a range of $[-25^{\circ}, +25^{\circ}]$ in out-of-plane angle relative to the current position. The out-of-plane interval is discretized in steps of $5^{\circ}$ which leads to 11 values to be predicted from each input image. 
For training, two different datasets were generated by forward projecting 3D volumes using the open-source physics-based x-ray simulator DeepDRR \cite{unberath2018,unberath2019enabling}. The resulting digitally reconstructed radiographs (DRRs) were created on a uniform grid with step size $5^{\circ}$ in both $\varphi$ and $\theta$. For each position $(\varphi, \theta)$, one clean image and one image with additional realistic noise injection were generated. The former was used to calculate ground truth detectability for each projection using equation~\ref{eq:dsquared} and the corresponding 3D scan while the latter was used as actual network input during training. The first dataset is based on five publicly available chest CT scans from the Cancer Imaging Archive (TCIA)~\cite{clark2013cancer}. Screw positions were manually annotated in six different vertebrae per scan. During the generation of projection data, only one vertebral level was considered at a time and a titanium screw was virtually inserted at the annotated position into the corresponding anatomy. Additionally, the isocenter of the simulated C-arm was varied randomly between the different simulations. 212 simulations were performed on 30 different anatomical sites, each resulting in 1368 images on a $5^{\circ}$ grid with a whole rotation ($[0^{\circ}, 360^{\circ}]$) for the in-plane angle and an interval of $[45^{\circ}, 135^{\circ}]$ for the out-of-plane angle. The resulting images of one chest CT scan are held out as a test set. Data generation for the second dataset is identical to the first dataset, but based on a semianthropomorphic representation of a human chest that is composed of a long box-like object, two cylinders, and two screws. The position of these objects was randomly varied within reasonable bounds to account for different anatomy from patient to patient finally leading to 75 simulations again consisting of 1368 images each, distributed over the same interval as above. The test set consists of three simulations.  
The first dataset is used for the experiments on synthetic data, while the second one is used to train the network in the real data case. Additionally, batch normalization and data augmentation using random rotations were included in the network for the real data experiments as we observe that it helps to improve generalization. 

\section{Experiments and Results}
\subsection{Simulation Experiments}
\label{sec:simulationexperiments}
\begin{figure}
    \centering
    \captionsetup[subfigure]{justification=centering}
    \begin{subfigure}{0.49\textwidth}
    	\centering
        \includegraphics[trim= 20 100 20 100,clip,width=\linewidth]{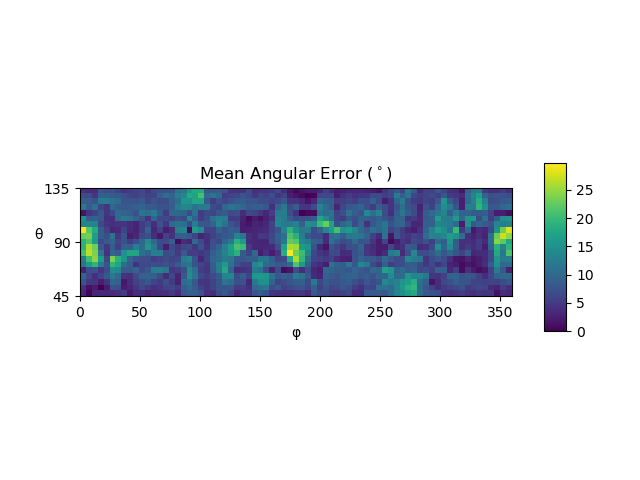}
        \caption{Angular error.}        
	\end{subfigure} 
	\begin{subfigure}{0.49\textwidth}
    	\centering
        \includegraphics[trim= 20 100 20 100,clip,width=\linewidth]{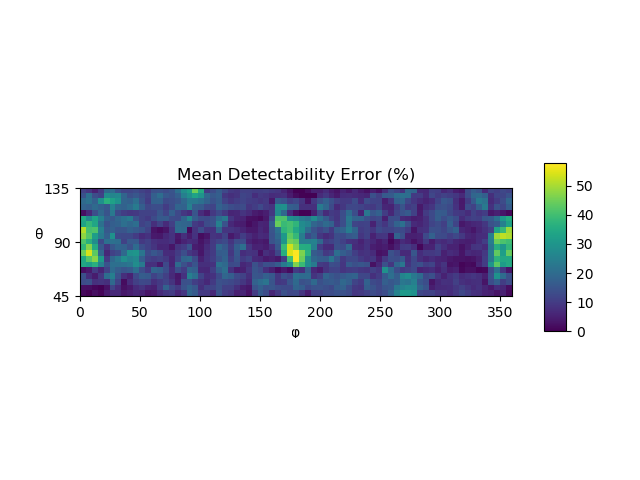}
        \caption{Detectability error.}        
	\end{subfigure}     
    \caption{Spatial distribution of the angular and detectability error. The X-axis shows the full $360^{\circ}$ in in-plane angle $\varphi$ and the Y-axis possible out-of-plane angles $\theta$ between $45^{\circ}$ and $135^{\circ}$}
    \label{fig:angular_distribution}
\end{figure}

\begin{figure}
    \centering
    \captionsetup[subfigure]{justification=centering}
    \begin{subfigure}{.22\textwidth}
          \centering
          % include first image
          \includegraphics[width = \textwidth]{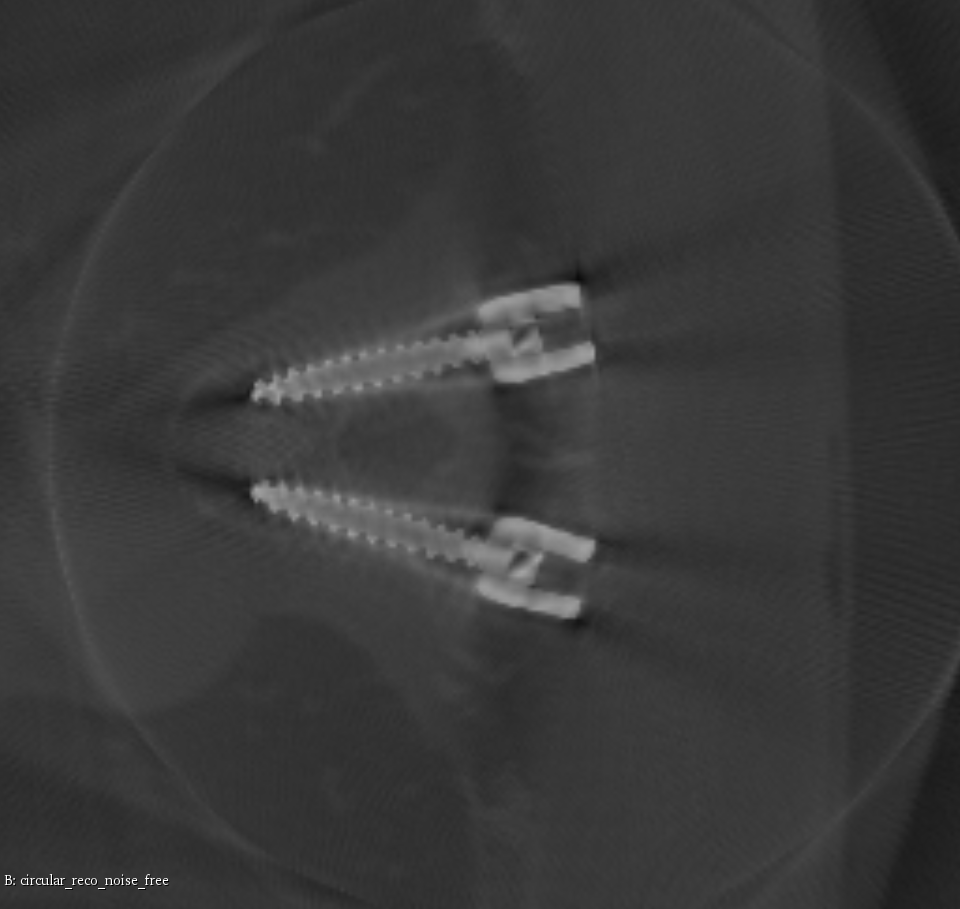}
    \end{subfigure}
    \begin{subfigure}{.22\textwidth}
          \centering
          % include second image
          \includegraphics[width = \textwidth]{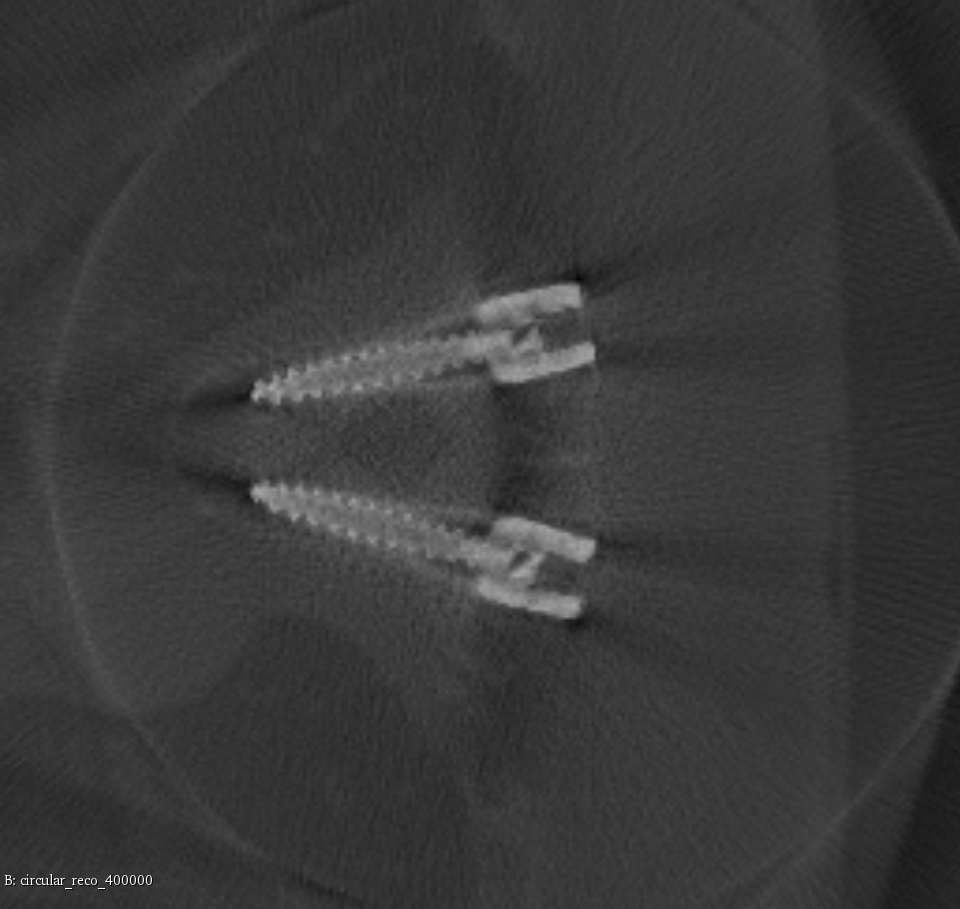}
    \end{subfigure}
    \begin{subfigure}{.22\textwidth}
          \centering
          % include second image
          \includegraphics[width = \textwidth]{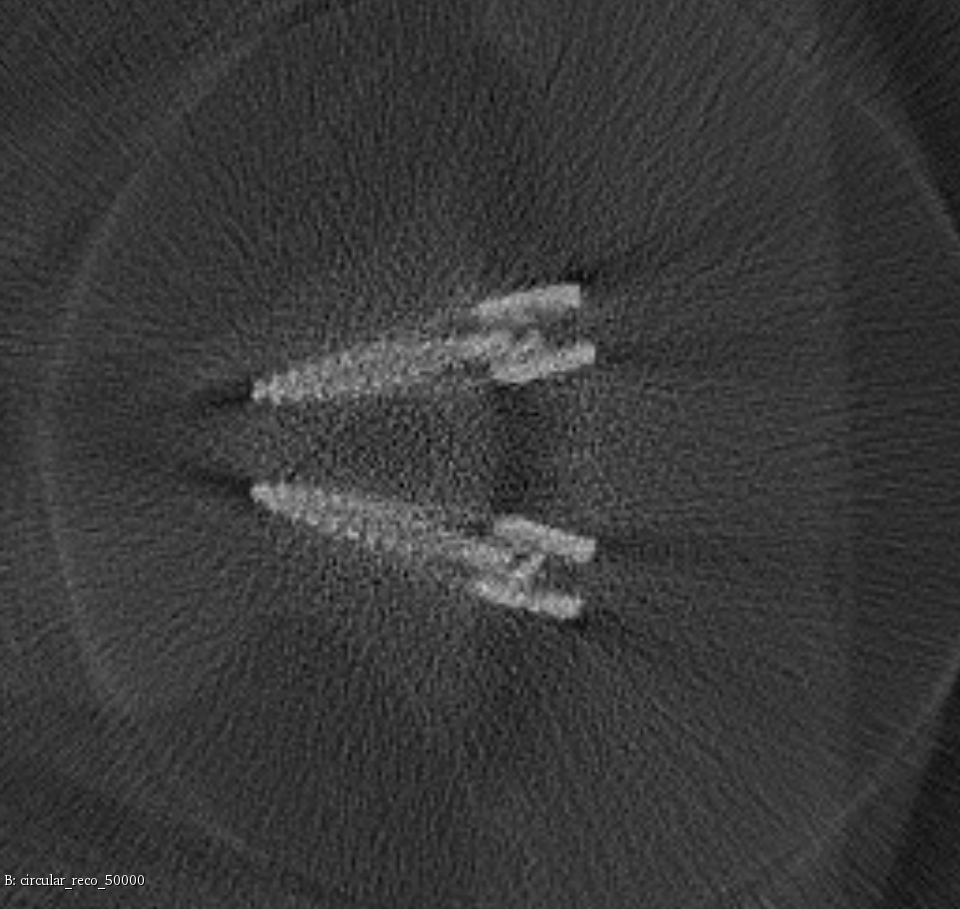}
    \end{subfigure}
    \begin{subfigure}{.22\textwidth}
          \centering
          % include second image
          \includegraphics[width = \textwidth]{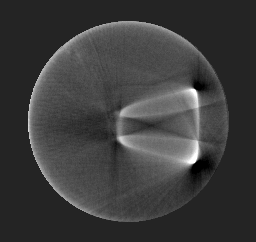}
    \end{subfigure}    
    
    \begin{subfigure}[t]{0.22\textwidth}
          \centering
          % include first image
          \includegraphics[width = \textwidth]{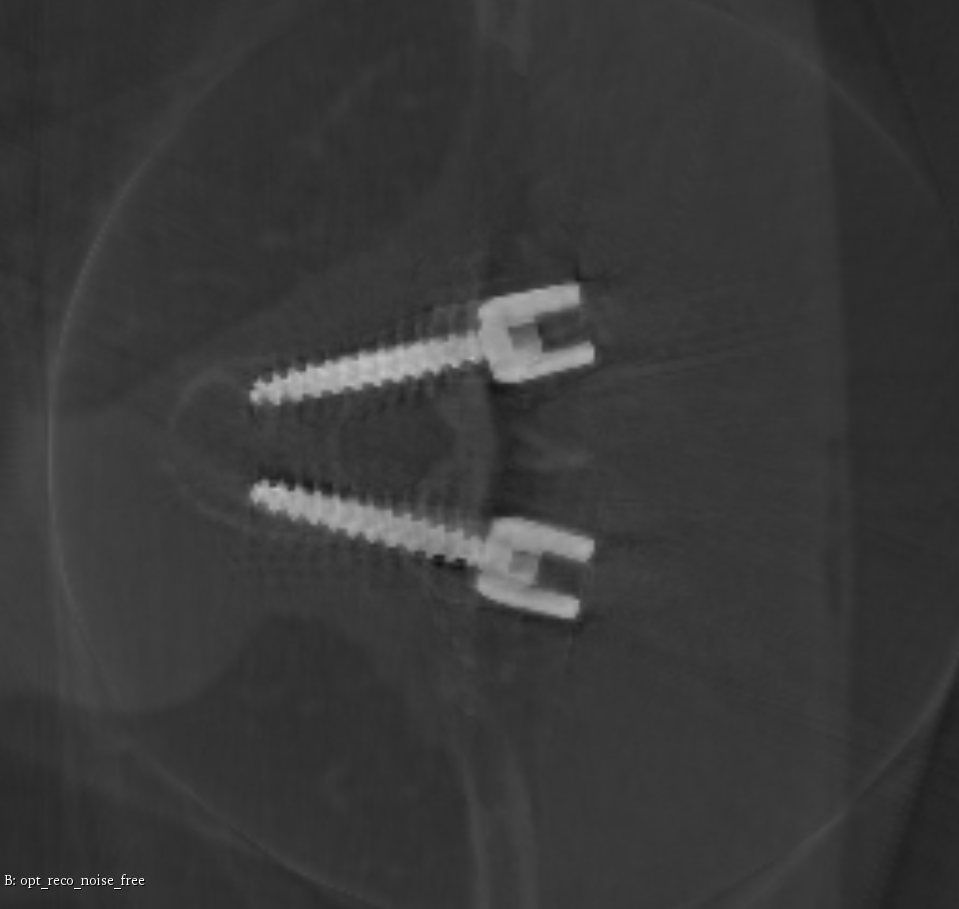}
          \caption{Simulation\newline no noise}
    \end{subfigure}
    \begin{subfigure}[t]{0.22\textwidth}
          \centering
          % include second image
          \includegraphics[width = \textwidth]{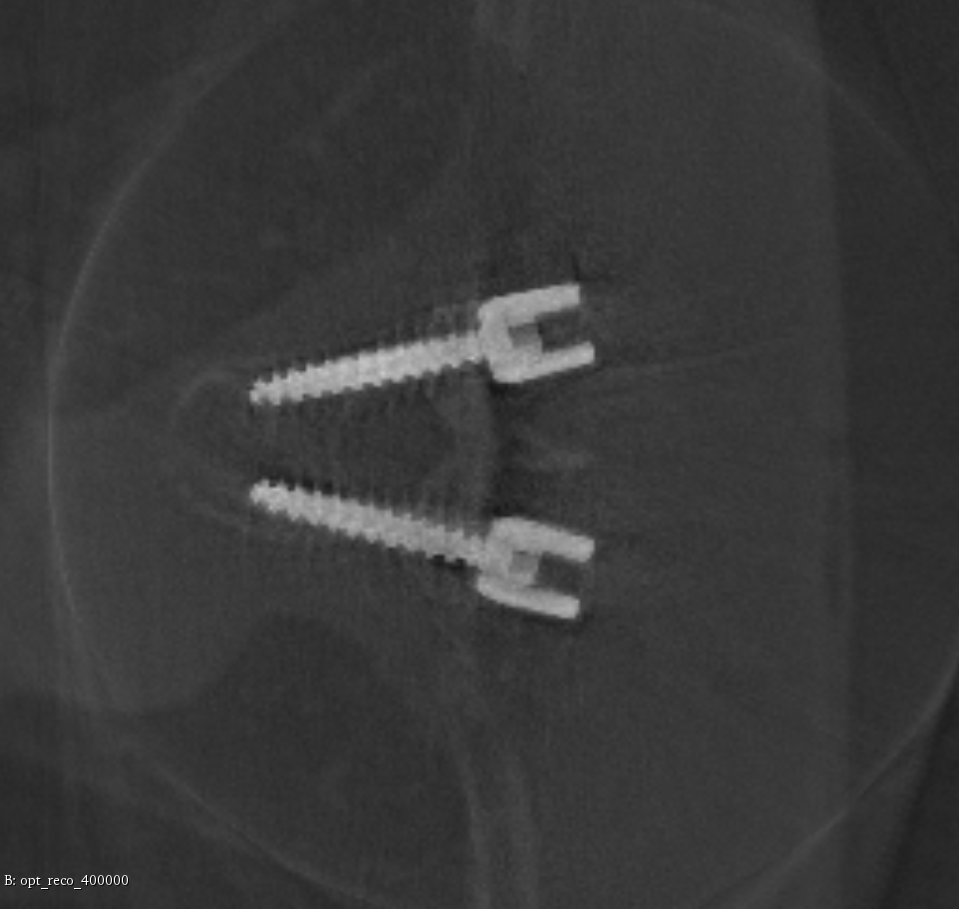}
          \caption{Simulation\newline $4\cdot10^5$ photons/pixel}
    \end{subfigure}
    \begin{subfigure}[t]{0.22\textwidth}
          \centering
          % include second image
          \includegraphics[width = \textwidth]{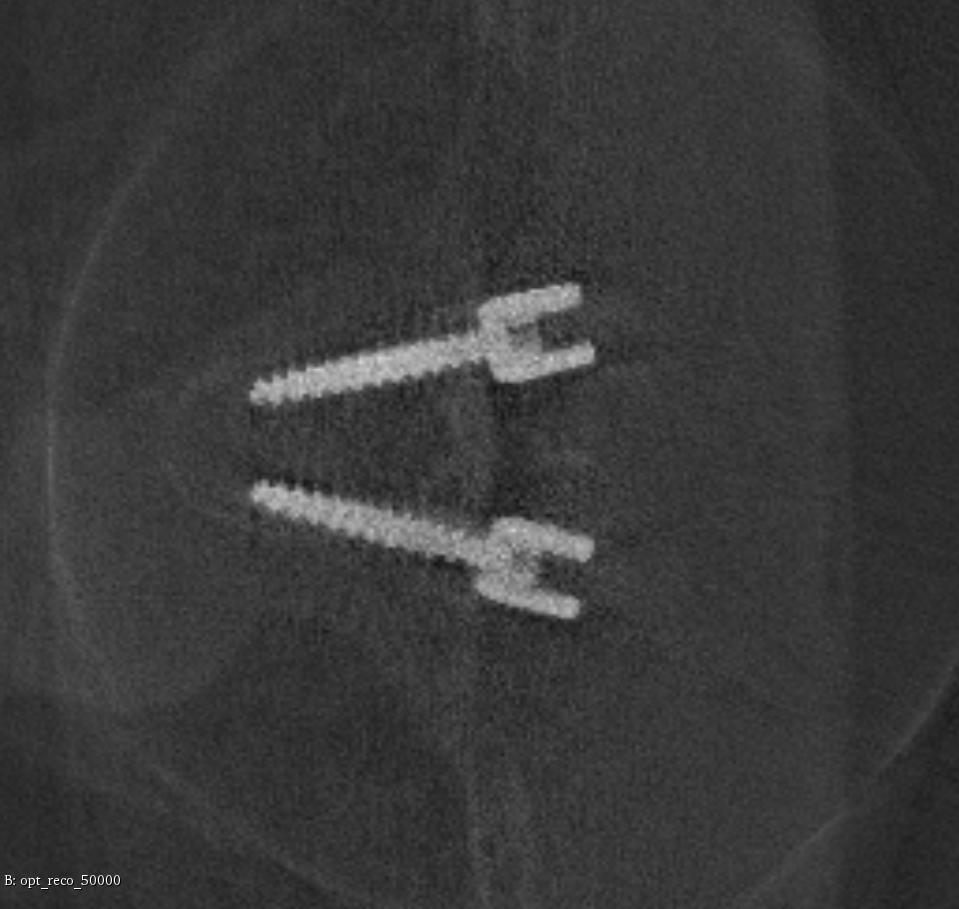}
          \caption{Simulation\newline $5\cdot10^4$ photons/pixel}
    \end{subfigure}
    \begin{subfigure}[t]{0.22\textwidth}
          \centering
          % include second image
          \includegraphics[width = \textwidth]{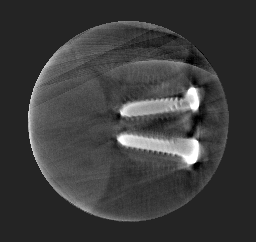}
          \caption{Real data}
    \end{subfigure}
    \caption{Slices through the reconstructions of synthetic and real data from a circular scan (upper row) and the task-aware trajectory (lower row) at different noise levels. Note that the simulated screws are not identical to the screws of the real phantom in size or shape.}
    \label{fig:reco}
\end{figure}

The network was first trained on the TCIA chest dataset. The training objective was to predict the detectabilities of the 11 projections with an offset of $+5^{\circ}$ in in-plane angle and an interval of $[-25^{\circ}, +25^{\circ}]$ in out-of-plane angle relative to the position of the input projection discretized in steps of $5^{\circ}$. During inference, an out-of-plane increment was chosen to be a step towards the highest predicted detectability. Additionally, the whole trajectory was restricted to an interval of $[-45^{\circ}, +45^{\circ}]$ concerning the out-of-plane angle relative to the starting position. 
%To assess the quality of the predicted trajectories based on the known ground truth detectability values of the test set, one can compute the angular distance between the best next out-of-plane angle predicted by the network and the corresponding angle selected using the ground truth information. This measure evaluates to $8.35^{\circ} \pm 11.61^{\circ}$. However, trajectories with high angular deviation from the optimal ground truth can still lead to nearly optimal reconstruction quality as the function of detectability values can be multimodal. To account for that, it is reasonable to also investigate the relative difference in detectability values between both the predicted and the optimal ground truth trajectory which is $13.69\% \pm 18.92\%$. 
\revised{In a purely simulated environment without realistic noise injection, the algorithm achieves $8.35^{\circ} \pm 11.61^{\circ}$ angular distance and $13.69\% \pm 18.92\%$ relative difference in detectability of the predicted trajectory compared to the ground truth \cite{zaech2019}. Their angular distributions are shown in Figure \ref{fig:angular_distribution}. In the following, the influence of different levels of noise and varying initialization poses of the C-arm on the prediction quality will be analyzed.}
%To investigate the influence of measurement noise on the predicted out-of-plane angle, 
Eight different $200^{\circ}$ short scan protocols are simulated for each screw pair in the test-set. Half of the protocols employ a circular trajectory, each with 200\,x-ray projections in total. These serve as the baseline protocols. The other half of the scans are generated on trajectories optimized with the proposed pipeline. For each of the two trajectory types, scans without noise and with a noise level corresponding to $5\cdot10^4$, $1\cdot10^5$, and $4\cdot10^5$\,photons per pixel are generated. Each x-ray projection is acquired with $620\times480$\,pixels and a pixel-size of $0.31\,\si{\milli\meter}\times 0.31\,\si{\milli\meter}$. This image corresponds to the central part of a standard flat-panel detector in $4\times4$ binning mode. A figure showing the predicted trajectories in the presence of different noise levels can be found in the supplementary material. Only small angular changes are observed which proves robustness against noise. Furthermore, the network did not overfit to a single detectability map, as the trajectories generated from different vertebral levels show major differences. Defining the trajectory in the noise-free case as the ground truth prediction allows calculation of the sensitivity to noise. The sensitivity is calculated as the angular mismatch, averaged over all angles and trajectories for a single noise level. For $4\cdot10^5$, $1\cdot10^5$, and $5\cdot10^4$\,photons per pixel, the mean angular error reads $0.83^{\circ}$, $1.13^{\circ}$, and $1.64^{\circ}$ respectively. The standard deviation of the predictions is $1.56^{\circ}$, $1.63^{\circ}$, and $1.73^{\circ}$ in the same order.
Besides robustness against noise, it is desirable that the optimal trajectory is largely independent of the starting angle. This property holds for the proposed algorithm, as a prediction only depends on the last acquired image. Therefore, two trajectories that intersect at any point will merge and continue as the same trajectory, given the noise is identical.
To show this property on data, the trajectories predicted from different starting angles, but the same anatomy were simulated (see plot in supplementary material). After few angle increments, the trajectories merge into two main bands that represent local maxima, which then merge into a single trajectory at $\phi = 50^{\circ}$. The initial differences of the trajectories can be explained by the limitations of the slope.
The predicted trajectories were reconstructed using a GPU implementation of the iterative conjugated gradient least squares algorithm for cone-beam geometry provided by the ASTRA toolbox \cite{vanaarle2015, vanaarle2016}. Figure~\ref{fig:reco} shows axial slices through the reconstructions from projections at different noise levels \revised{for qualitative analysis. For quantitative assessment, comparison is performed by computing the full width half maximum (FWHM) of the screws of one vertebral level averaged over two different positions which quantifies the amount of blooming artifact. Further, we investigate the intensity of the Fourier spectrum of a small normalized image patch containing the screw thread at the frequency of the thread itself. For comparison, the ground truth value for each of these measures is listed which is obtained by reconstructing mono-energetic, noise-free simulated projections without any physics-based artifacts. We also report the structural similarity (SSIM) of a slice containing both screws between the ground truth and the noisy reconstructions. Results are reported in table~\ref{tab:results_sim}. Both the FWHM and the thread frequency height are closer to the true value for the task-aware trajectories compared to the circular ones. Also the image slices extracted from the reconstructions are more similar to the ground truth slices as indicated by higher SSIM values. Noise in general deteriorates the reconstruction performance, but this seems to be less severe for the task-aware trajectories.}
\begin{table}
    \centering
    \begin{tabular}{c|c|c|c}
         & screw FWHM [mm] & thread frequency height & SSIM to ground truth\\ \hline
         ground truth & \revised{3.92} & \revised{6.83} & \revised{1.00} \\
         \hline
         circular no noise & \revised{6.38} & \revised{9.05} & \revised{0.83} \\ 
         circular $4\cdot10^5$ photons/pixel & \revised{6.35} & \revised{8.96} & \revised{0.81}\\ 
         circular $5\cdot10^4$ photons/pixel & \revised{6.19} & \revised{7.52} & \revised{0.69}\\ \hline
         task-aware no noise & \revised{3.65} & \revised{7.57} & \revised{0.90}\\
         task-aware $4\cdot10^5$ & \revised{3.72} & \revised{8.31} & \revised{0.89}\\
         task-aware $5\cdot10^4$ & \revised{4.10} & \revised{6.95} & \revised{0.85}\\
    \end{tabular}
    \caption{\revised{Evaluation of reconstruction quality based on screw FWHM, screw thread frequency peak height and SSIM for circular and task-aware trajectories and different noise levels on simulated data.}}
    \label{tab:results_sim}
\end{table}

\subsection{Real Data Experiments}
\revised{One central challenge when implementing non-circular orbits on any CBCT scanner - robotized or conventional C-arm - is the calibration required for precise reconstruction. Usually only few reproducible circular short-scan trajectories are calibrated in advance using phantoms specifically designed for this purpose. The trajectories aimed for here, however, cannot be precalibrated because they are scene specific, and thus, not known in advance. To overcome this challenge, in this work, multiple pre-calibrated CBCT short-scans of the phantom were acquired at various swivel and tilt angles prior to trajectory prediction. The CBCT reconstructions, and via pre-calibration also all projection images, acquired in this way were then aligned in a common 3D object space using image-based registration. This procedure aims at providing a sufficient sampling of all possible views $(\varphi, \theta)$ and is explained in detail in the following paragraph. During inference, the sampled view closest to the predicted optimal view is identified and used as subsequent input for the network instead of an image acquired in real-time by a robotic device. This allows to predict non-circular trajectories from real data using the proposed method in a retrospective manner and avoids the need for a fully robotic C-arm. Instead, data acquisition was performed on a conventional CBCT scanner (Siemens Arcadis Orbic 3D).}
For the experiments, a phantom was built in line with the simulated training data for this case. It consists of two screws drilled into a wooden rod and two cylinders filled with ballistic gel.
\begin{figure}
    \centering
    \captionsetup[subfigure]{justification=centering}
    \begin{subfigure}{0.375\textwidth}
    	\centering
        \includegraphics[width=\textwidth,angle=90]{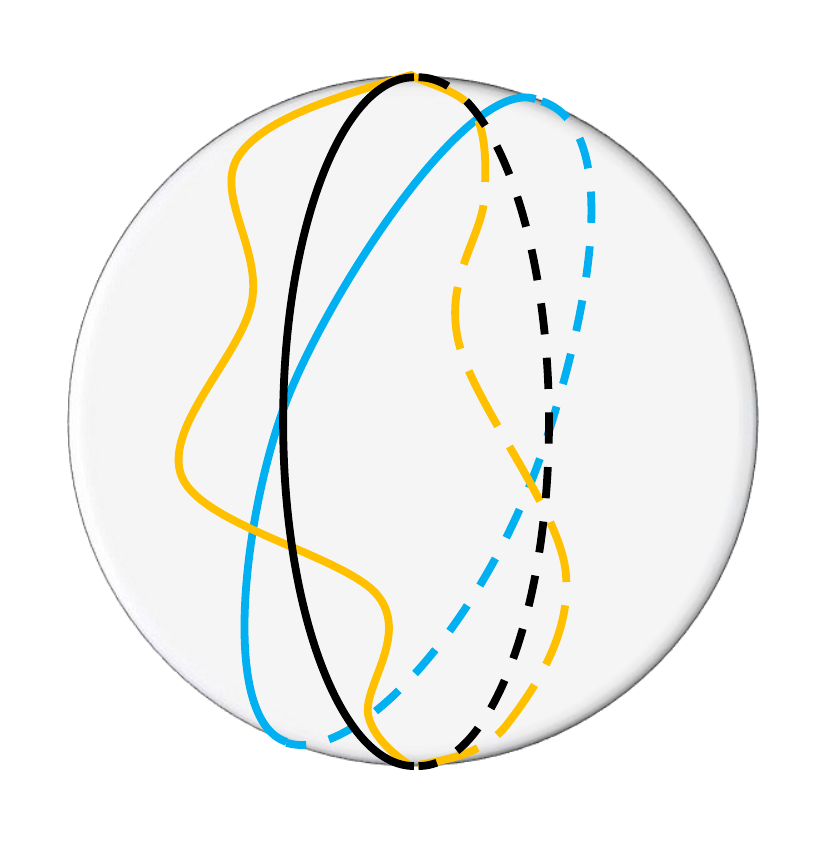}       
	\end{subfigure} 
	\begin{subfigure}{0.525\textwidth}
    	\centering
        \includegraphics[width=\textwidth]{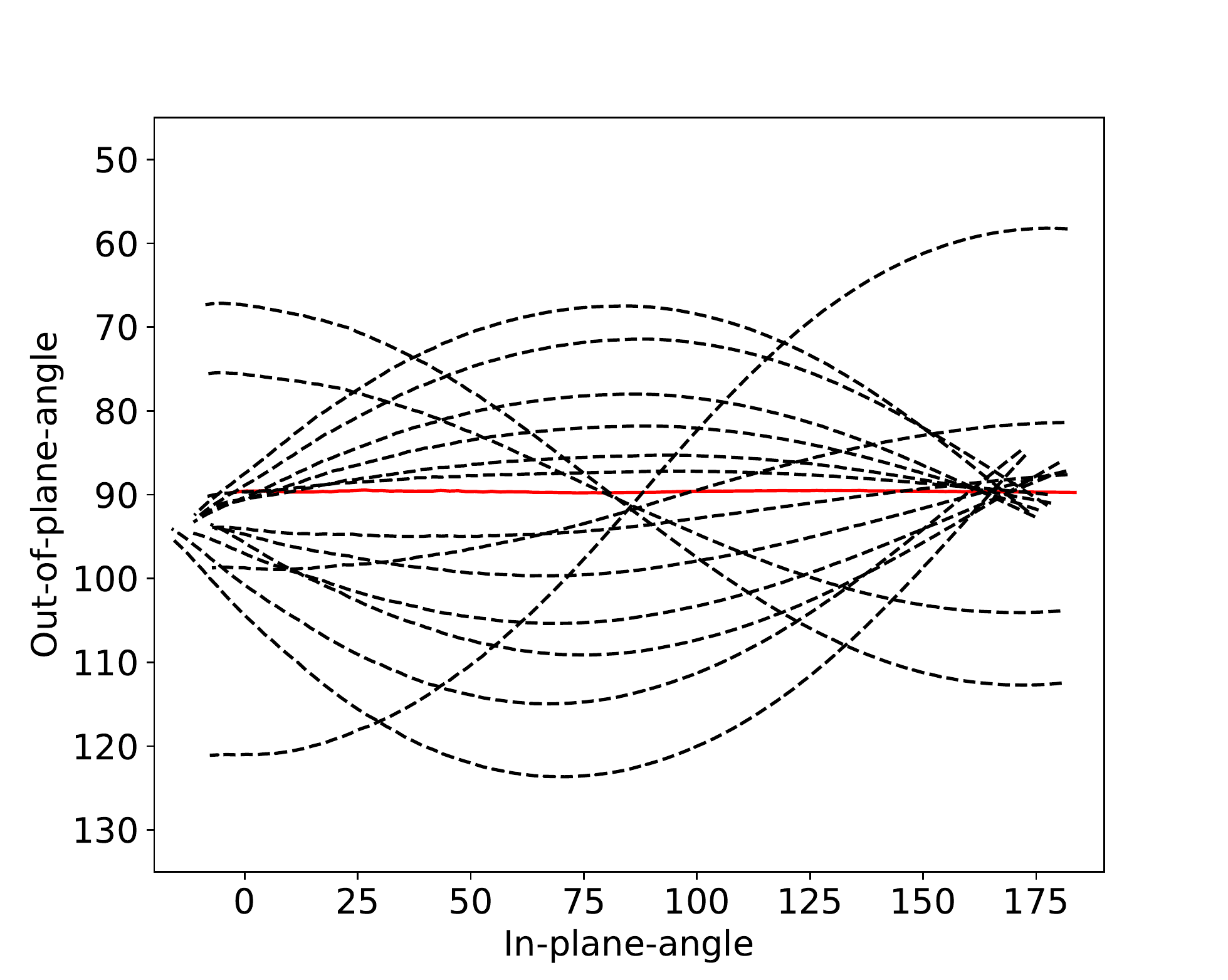}       
	\end{subfigure}     
     \caption{Left: Two exemplary tilted orbits and a non-circular trajectory with varying out-of-plane angle. Right: Sampling of the $(\varphi, \theta)$-space using tilted orbits. Solid red represents the untilted reference scan, dashed black refers to scans acquired on tilted circular orbits.}
    \label{fig:sampling}
\end{figure}

\begin{figure}
    \centering
    \includegraphics[trim= 3cm 0 1cm 0,clip,width=\textwidth]{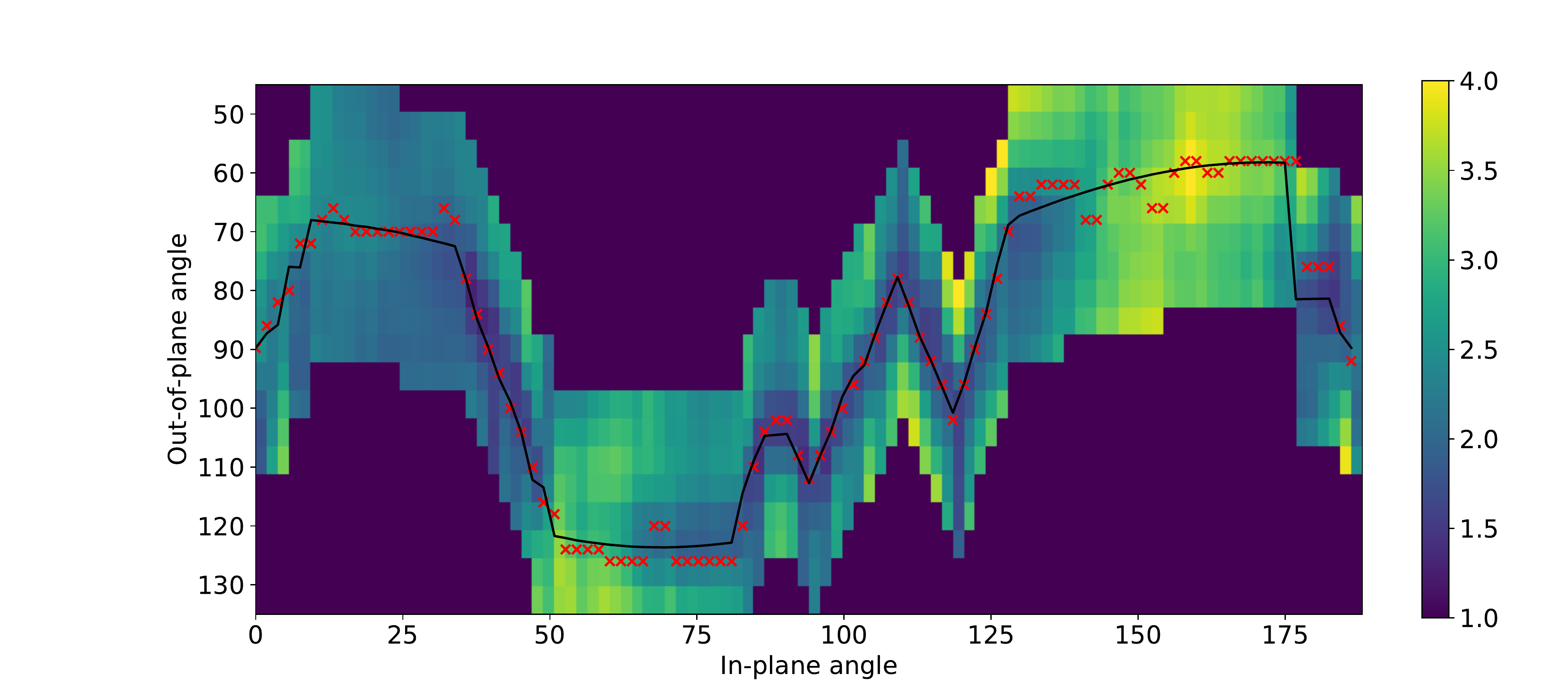}
    \caption{Predicted trajectory on real data in black and network predictions relative to the current position. The crosses show optimal views based on network output, the black line is the final trajectory based on the closest sampled view.}
    \label{fig:realtrajectory}
\end{figure}

Sampling the $(\varphi, \theta)$ space using only circular scans can be achieved by scanning the phantom on tilted but circular orbits (see figure~\ref{fig:sampling} left). However, tilting the scanner would require calibration of each of these tilted trajectories due to mechanical sagging and wobble. Instead, the position of the phantom itself was altered between successive scans while the scanner trajectory was kept identical. In this manner, 17 scans were acquired mimicking tilt as well as swivel of the C-arm. In terms of the in-plane and out-of-plane angle notation, each of these 17 scans results in a curve of sampled views in the $(\varphi, \theta)$ space (see figure~\ref{fig:sampling} right).
All 17 scans were reconstructed and 16 tilted scans were rigidly registered to one reference volume. Registration was performed by optimizing a normalized cross-correlation objective function using quadratic optimization (BOBYQA). The rigid transformation $T_i$ aligning the i-th moving tilted volume with the reference volume was obtained and used to adjust the projection matrices as:
\begin{equation}
    P_{tilt}^{i} = T_i^{-1} P_{flat}
\end{equation}
Applying the inverse transformation to the projection matrices allows changing from several volumes reconstructed with the same set of matrices $P_{flat}$ to a scan-specific set of matrices $P_{tilt}^{i}$ such that all projections can be integrated into the same volume during reconstruction. 
The network was trained on the dataset created from a digital copy of the used phantom mentioned in section~\ref{sec:network}. Training ground truth was chosen to be identical to the setup described for the synthetic data experiments in section~\ref{sec:simulationexperiments}. During inference, increments in in-plane angle were fixed to $\Delta\varphi = 1^{\circ}$ and the out-of-plane angle step is computed from the predicted detectability and a regularization component that penalizes high directional changes and promotes a smoother trajectory. As the pipeline is targeted to be implemented on a robotic C-arm device, we need to account for the limited mechanical capabilities of such a system. Sudden directional changes would require high accelerations that cannot be realized safely. Therefore, we introduce the cosine of the angle between two subsequent steps as additional penalty term. With this term, sudden directional changes are traded off with best next steps as predicted by the network.    
\begin{equation}
    \theta_{t+1} = \theta_{t} + \max_i(\lambda (u\cdot v_i) + p_i)
\end{equation}
Here, $u$ denotes the previous trajectory direction in terms of $(\Delta \varphi, \Delta \theta)$, $v_i$ is the i-th possible next direction and $p_i$ is the corresponding predicted detectability. We heuristically find that $\lambda = 0.6$ is a suitable weighting factor and keep it constant for all experiments. The projection image which is closest to the optimal predicted view in terms of $\varphi$ and $\theta$ is identified from the set of acquired projections, added to the trajectory, deleted from the set of available sampled views for all following steps, and used as next input for the network. Using the first projection of the reference scan which corresponds to an out-of-plane angle of $90^{\circ}$ (scan plane intersecting long axis of both screws) as initialization of the algorithm, this procedure results in the trajectory depicted in figure~\ref{fig:realtrajectory}.
From the initial out-of-plane angle, the algorithm proposes to increase the tilt of the C-arm for the majority of the scan. The trajectory reaches the most extreme sampled out-of-plane angles in positive direction for in-plane angles $50^{\circ}$ to $80^{\circ}$ and the most extreme angles in negative direction towards beginning and end of the scan. In the central part, it exhibits a slightly alternating behavior. Note that in the real data case, only views that have been sampled can be part of the trajectory which limits the number of possible solutions considerably. 
Reconstructions were calculated using the same algorithm as for the synthetic data \cite{vanaarle2015, vanaarle2016}. Projection images were masked prior to reconstruction based on forward projecting a centered sphere with \SI{5}{\centi\meter} radius in 3D to reduce truncation artifacts and the algorithm was executed for 300 iterations. A slice through the reconstructed volume of the trajectory corresponding to figure~\ref{fig:realtrajectory} and the circular reference trajectory can be found in the last column of figure~\ref{fig:reco}. While the overall shape of the two screws, as well as its threads, are only poorly recovered in the reconstruction from the circular trajectory, the task aware protocol is able to recover much finer structures.
For quantitative evaluation, we additionally initialize our algorithm with the first projection of the four swivel trajectories in our dataset, which each provide initialization with a different out-of-plane angle.
%and compute a trajectory based on a simple heuristic instead of the network predictions: In each projection image, the screws are segmented and a higher quality index is assigned, the more screw pixel have been found. This is based on the intuition that seeing a large cross sectional area of the screws implies a short path of the photons through the metal. 
We compare reconstructions obtained from all these trajectories to the reference circular trajectory and the two circular trajectories of our dataset associated with the highest tilt and swivel respectively. \revised{Comparison is again performed using the FWHM of the screw and the thread frequency peak height. Results can be found in table~\ref{tab:results}. Calculating the SSIM is not possible as no ground truth information is available.} 
\begin{table}
    \centering
    \begin{tabular}{c|c|c}
         & screw FWHM [mm] & thread frequency height \\ \hline
         circular reference & 12.67 & 2.11 \\ 
         circular max. tilt & 6.29 & 3.24 \\ 
         circular max.swivel & 9.36 & 9.79 \\ \hline
         %heuristic & 6.37 & 1.91\\ \hline
         task-aware init $67^{\circ}$ & 7.35 & 5.98\\
         task-aware init $76^{\circ}$ & 7.29 & 5.99 \\
         task-aware init $90^{\circ}$ & 6.93 & 4.33 \\
         task-aware init $99^{\circ}$ & 6.92 & 4.38 \\
         task-aware init $121^{\circ}$ & 6.79 & 9.42 \\
    \end{tabular}
    \caption{Evaluation of reconstruction quality based on screw FWHM and screw thread frequency peak height for different trajectories.}
    \label{tab:results}
\end{table}
The circular reference scan performs worst by far, exhibiting the largest FWHM of all scans and revealing severe problems in visualizing the shape of the screw. While the trajectories corresponding to maximum tilt and swivel perform best when considering either FWHM or peak height, respectively, the task-aware trajectories can improve both measures decisively compared to the reference scan. Initializing with angles different from the reference scan ($90^{\circ}$ out-of-plane) additionally seems to improve the ability to reconstruct the screw thread.

\section{Discussion}
The presented results on simulated data \revised{help to understand strengths and limitations of the method in a controlled setting and serve as an upper bound of the ideally achieved performance. They} show that predicting the detectability values of possible next views from the current projection is possible with reasonable accuracy and robustness against different noise levels and initialization angles. The resulting trajectories are in line with previously published concepts on the emergence of reconstruction artifacts introduced in section~\ref{sec:intro}. If possible, our algorithm avoids views with overlapping screws as well as views along the screws' long axes, which cause the most severe inconsistencies (beam hardening up to photon starvation) based on the assumptions made during reconstruction. Therefore, the fine structures of the screws can be reconstructed with higher quality and metal artifacts can be reduced significantly. 
The performed real data experiments hint at the feasibility of the approach in real CBCT acquisitions. \revised{Benchmarks for the inference time of VGG-19 point out that it is generally feasible to use the network predictions for real-time adjustment of the C-arm \cite{canziani2016}, but as our real data evaluation was performed retrospectively, we did neither implement a real-time capable system including the mechanical components nor did we investigate whether the final trajectory can be realized by a scanner within reasonable scanning time. The retrospective evaluation still suggests that} the task-aware trajectories lead to considerably improved reconstructions of the screws' general shape as well as its thread. This holds true especially when comparing to the reference scan, the scan plane of which is parallel to both screws' long axes. Unfortunately, this acquisition scheme is most predominantly employed in the operating room. This leads to the conclusion that slightly re-positioning the C-arm to acquire a short scan trajectory that is tilted with respect to the standard plane already avoids many of the worst views, and would thus already result in considerably improved reconstruction quality \emph{without changes} to the routine acquisition protocol.

On real data, the predicted trajectory shows an increased alternating behavior between positive and negative out-of-plane angles compared to the simulations. \revised{Possible reasons for this are a sub-optimal generalization of the network from its training domain to the domain of real images which could potentially be mitigated by an increased amount of training data.} Moreover, the network fails to disambiguate positive and negative increments for the out-of-plane angle in some cases while still clearly following the trend to favor high out-of-plane angles in our specific setup. This behavior results in trajectories that tend to jump between high positive and high negative out-of-plane angles and might be caused by the Markov property of the algorithm described here. As each prediction is only based on one preceding projection image, there is very little contextual information available that could be used to disambiguate predictions with similar detectability. The predicted trajectory still leads to remarkable improvements in the reconstruction results compared to the circular reference trajectory which can already improve the ability for accurate clinical assessment tasks. 
\revised{Still, there are many open challenges which need to be addressed to push the approach closer to the level of accuracy and robustness needed for clinical application. First, the retrospective calibration procedure presented here on a non-robotic C-arm with only one actuated axis to enable CBCT is not applicable in a clinical setting because it would expose the patient to high doses of ionizing radiation. Instead, an online calibration procedure which does not require precise knowledge about the 3D structure would be desirable. Relying on the joint encodings of a fully robotic C-arm for initialization, further fine-tuning of the pose parameters could be performed in an image-based manner, e.g. using autofocus measures \cite{preuhs2020}. To ensure robust and precise network predictions in a clinical environment, important steps are the generation of synthetic training data, which is representative of the variety of different anatomies and tools as well as views onto these. The domain gap between the simulations used for training and the real fluoroscopy images during inference could be minimized using state-of-the-art domain adaptation techniques. Once a task-aware protocol is deployed in practice and a broader spectrum of fluoroscopy images from different views onto the anatomy becomes available, real data from predicted trajectories can be used to directly retrain the network parameters. Experimenting with different network architectures might also improve prediction performance. Finally, we envision a clinically applicable version of the pipeline to supersede existing CBCT protocols as they are already applied in the operating room.}

\section{Conclusion}
We introduced a learning-based method for online CBCT trajectory adjustment that overcomes the need for volumetric information at imaging time. This is the first step towards high-quality intra-operative C-arm CBCT imaging which is based on the idea of directly acquiring better data for artifact avoidance. Such an approach might ultimately enable intraoperative verification of implant placement with high confidence, as is required for high volume procedures including spinal fusion surgery. Future work will address the lack of sequential modeling in the current approach and investigate whether the image quality delivered by a refined version of our approach is sufficient for clinical interpretation.  

\textbf{Acknowledgements} We gratefully acknowledge the support of ~R01 EB023939, ~R01 EB016703, ~R21 EB028505, and JHU Internal Funds. We further acknowledge the support of the NVIDIA Corporation with the donation of the GPUs used for this research. We would like to thank Gerhard Kleinszig and Sebastian Vogt, both with Siemens Healthineers, for making the C-arm used in this research available. Mareike Thies was supported by a DAAD Promos Fellowship.

~\\
\textbf{Disclaimer:} The concepts and information presented in this paper are based
on research and are not commercially available.

~\\
\textbf{Conflict of interest:} The authors have no conflict of interest to declare. 

~\\
\textbf{Ethical approval:} This article does not describe research on human subjects, therefore, ethical approval was not required.

~\\
\textbf{Informed consent:} This article does not contain patient data.

\bibliographystyle{plain}
\bibliography{bibliography}   % name your BibTeX data base

\begin{thebibliography}{10}

\bibitem{szpalski2010}
Max Aebi.
\newblock {I}ndication for {L}umbar {S}pinal {F}usion.
\newblock In Marek Szpalski, Robert Gunzburg, Björn~L. Rydevik, Jean-Charles
  Le~Huec, and H.~Michael Mayer, editors, {\em Surgery for Low Back Pain},
  pages 109--122. Springer, 2010.

\bibitem{andersson2014}
Gunnar Andersson and Sylvia~I Watkins-Castillo.
\newblock {S}pinal {F}usion.
\newblock {\em The Burden of Musculoskeletal Diseases in the United States
  (BMUS)}, 2014.
\newblock Available at \url{https://www.boneandjointburden.org/}, accessed
  02/17/2020.

\bibitem{canziani2016}
Alfredo Canziani, Adam Paszke, and Eugenio Culurciello.
\newblock {A}n analysis of deep neural network models for practical
  applications.
\newblock {\em arXiv preprint arXiv:1605.07678}, 2016.

\bibitem{clark2013cancer}
Kenneth Clark, Bruce Vendt, Kirk Smith, John Freymann, Justin Kirby, Paul
  Koppel, Stephen Moore, Stanley Phillips, David Maffitt, Michael Pringle, and
  L~Tarbox.
\newblock The {C}ancer {I}maging {A}rchive ({TCIA}): {M}aintaining and
  operating a public information repository.
\newblock {\em Journal of digital imaging}, 26(6):1045--1057, 2013.

\bibitem{cordemans2017}
Virginie Cordemans, Ludovic Kaminski, Xavier Banse, Bernard~G Francq, and
  Olivier Cartiaux.
\newblock {A}ccuracy of a new intraoperative cone beam {CT} imaging technique
  ({A}rtis zeego {II}) compared to postoperative {CT} scan for assessment of
  pedicle screws placement and breaches detection.
\newblock {\em European Spine Journal}, 26(11):2906--2916, 2017.

\bibitem{dabravolski2014}
Andrei Dabravolski, Kees~Joost Batenburg, and Jan Sijbers.
\newblock {D}ynamic angle selection in x-ray computed tomography.
\newblock {\em Nuclear Instruments and Methods in Physics Research Section B:
  Beam Interactions with Materials and Atoms}, 324:17--24, 2014.

\bibitem{deyo2004}
Richard~A Deyo, Alf Nachemson, and Sohail~K Mirza.
\newblock {S}pinal-fusion surgery -- the case for restraint.
\newblock {\em The Spine Journal}, 4(5):138--142, 2004.

\bibitem{feldkamp1984}
Lee~A Feldkamp, Lloyd~C Davis, and James~W Kress.
\newblock {P}ractical cone-beam algorithm.
\newblock {\em Josa a}, 1(6):612--619, 1984.

\bibitem{fritzell2003}
Peter Fritzell, Olle H{\"a}gg, and Anders Nordwall.
\newblock {C}omplications in lumbar fusion surgery for chronic low back pain:
  {C}omparison of three surgical techniques used in a prospective randomized
  study. {A} report from the {S}wedish {L}umbar {S}pine {S}tudy {G}roup.
\newblock {\em European spine journal}, 12(2):178--189, 2003.

\bibitem{fu2008}
Tsai-Sheng Fu, Chak-Bor Wong, Tsung-Ting Tsai, Yen-Chiu Liang, Lih-Huei Chen,
  and Wen-Jer Chen.
\newblock {P}edicle screw insertion: {C}omputed tomography versus fluoroscopic
  image guidance.
\newblock {\em International orthopaedics}, 32(4):517--521, 2008.

\bibitem{gang2020}
Grace~J. Gang, Jeffrey~H. Siewerdsen, and J.~Webster Stayman.
\newblock {N}on-circular {CT} orbit design for elimination of metal artifacts.
\newblock In Guang-Hong Chen and Hilde Bosmans, editors, {\em Medical Imaging
  2020: Physics of Medical Imaging}, volume 11312, pages 531 -- 536.
  International Society for Optics and Photonics, SPIE, 2020.

\bibitem{gang2014}
Grace~J Gang, J~Webster Stayman, Wojciech Zbijewski, and Jeffrey~H Siewerdsen.
\newblock {T}ask-based detectability in {CT} image reconstruction by filtered
  backprojection and penalized likelihood estimation.
\newblock {\em Medical physics}, 41(8Part1):081902, 2014.

\bibitem{garber2012}
Sarah~T Garber, Erica~F Bisson, and Meic~H Schmidt.
\newblock {C}omparison of three-dimensional fluoroscopy versus postoperative
  computed tomography for the assessment of accurate screw placement after
  instrumented spine surgery.
\newblock {\em Global spine journal}, 2(2):095--098, 2012.

\bibitem{gelalis2012accuracy}
Ioannis~D Gelalis, Nikolaos~K Paschos, Emilios~E Pakos, Angelos~N Politis,
  Christina~M Arnaoutoglou, Athanasios~C Karageorgos, Avraam Ploumis, and
  Theodoros~A Xenakis.
\newblock Accuracy of pedicle screw placement: {A} systematic review of
  prospective in vivo studies comparing free hand, fluoroscopy guidance and
  navigation techniques.
\newblock {\em European Spine Journal}, 21(2):247--255, 2012.

\bibitem{kondo2010}
Atsushi Kondo, Yoshihiko Hayakawa, Jian Dong, and Akira Honda.
\newblock {I}terative correction applied to streak artifact reduction in an
  {X}-ray computed tomography image of the dento-alveolar region.
\newblock {\em Oral radiology}, 26(1):61--65, 2010.

\bibitem{liao2019}
Haofu Liao, Wei-An Lin, Zhimin Huo, Levon Vogelsang, William~J Sehnert, S~Kevin
  Zhou, and Jiebo Luo.
\newblock {G}enerative {M}ask {P}yramid {N}etwork for {CT}/{CBCT} {M}etal
  {A}rtifact {R}eduction with {J}oint {P}rojection-{S}inogram {C}orrection.
\newblock In {\em International Conference on Medical Image Computing and
  Computer-Assisted Intervention}, pages 77--85. Springer, 2019.

\bibitem{manbachi2016}
Amir Manbachi.
\newblock {\em {T}owards {U}ltrasound-guided {S}pinal {F}usion {S}urgery}.
\newblock Springer, 2016.

\bibitem{meilinger2011}
Manuel Meilinger, Christian Schmidgunst, Oliver Sch{\"u}tz, and Elmar~W Lang.
\newblock {M}etal artifact reduction in cone beam computed tomography using
  forward projected reconstruction information.
\newblock {\em Zeitschrift f{\"u}r medizinische Physik}, 21(3):174--182, 2011.

\bibitem{milletari2019}
Fausto Milletari, Vighnesh Birodkar, and Michal Sofka.
\newblock {S}traight to the point: {R}einforcement learning for user guidance
  in ultrasound.
\newblock In {\em Smart Ultrasound Imaging and Perinatal, Preterm and
  Paediatric Image Analysis}, pages 3--10. Springer, 2019.

\bibitem{preuhs2020}
Alexander Preuhs, Michael Manhart, Philipp Roser, Bernhard Stimpel, Christopher
  Syben, Marios Psychogios, Markus Kowarschik, and Andreas Maier.
\newblock {D}eep autofocus with cone-beam {CT} consistency constraint.
\newblock In {\em Bildverarbeitung f{\"u}r die Medizin 2020}, pages 169--174.
  Springer, 2020.

\bibitem{schulze2011}
R~Schulze, U~Heil, D~Gro{\ss}, DD~Bruellmann, E~Dranischnikow, U~Schwanecke,
  and E~Schoemer.
\newblock {A}rtefacts in {CBCT}: {A} review.
\newblock {\em Dentomaxillofacial Radiology}, 40(5):265--273, 2011.

\bibitem{schulze2010}
Ralf Kurt~Willy Schulze, Dorothea Berndt, and Bernd d'Hoedt.
\newblock {O}n cone-beam computed tomography artifacts induced by titanium
  implants.
\newblock {\em Clinical oral implants research}, 21(1):100--107, 2010.

\bibitem{sembrano2012}
Jonathan~N Sembrano, David~W Polly, Charles Gerald~T Ledonio, and Edward
  Rainier~G Santos.
\newblock {I}ntraoperative 3-dimensional imaging ({O}-arm) for assessment of
  pedicle screw position: {D}oes it prevent unacceptable screw placement?
\newblock {\em International journal of spine surgery}, 6:49--54, 2012.

\bibitem{simonyan2014}
Karen Simonyan and Andrew Zisserman.
\newblock {V}ery deep convolutional networks for large-scale image recognition.
\newblock {\em arXiv preprint arXiv:1409.1556}, 2014.

\bibitem{stayman2013}
J~Webster Stayman and Jeffrey~H Siewerdsen.
\newblock {T}ask-based trajectories in iteratively reconstructed interventional
  cone-beam {CT}.
\newblock {\em Proc. 12th Int. Meet. Fully Three-Dimensional Image Reconstr.
  Radiol. Nucl. Med}, pages 257--260, 2013.

\bibitem{thomsen1997}
K~Thomsen, Finn~B Christensen, S{\o}ren~P Eiskj{\ae}r, Ebbe~S Hansen, S{\o}ren
  Fruensgaard, and Cody~E. B{\"u}nger.
\newblock {T}he effect of pedicle screw instrumentation on functional outcome
  and fusion rates in posterolateral lumbar spinal fusion: {A} prospective
  randomized clinical study.
\newblock {\em Spine}, 22(24):2813--2822, 1997.

\bibitem{unberath2019enabling}
Mathias Unberath, Jan-Nico Zaech, Cong Gao, Bastian Bier, Florian Goldmann,
  Sing~Chun Lee, Javad Fotouhi, Russell Taylor, Mehran Armand, and Nassir
  Navab.
\newblock {E}nabling machine learning in {X}-ray-based procedures via realistic
  simulation of image formation.
\newblock {\em International journal of computer assisted radiology and
  surgery}, 14(9):1517--1528, 2019.

\bibitem{unberath2018}
Mathias Unberath, Jan-Nico Zaech, Sing~Chun Lee, Bastian Bier, Javad Fotouhi,
  Mehran Armand, and Nassir Navab.
\newblock {D}eep{DRR} -- a catalyst for machine learning in fluoroscopy-guided
  procedures.
\newblock In {\em International Conference on Medical Image Computing and
  Computer-Assisted Intervention}, pages 98--106. Springer, 2018.

\bibitem{vanaarle2016}
Wim Van~Aarle, Willem~Jan Palenstijn, Jeroen Cant, Eline Janssens, Folkert
  Bleichrodt, Andrei Dabravolski, Jan De~Beenhouwer, K~Joost Batenburg, and Jan
  Sijbers.
\newblock {F}ast and flexible {X}-ray tomography using the {ASTRA} toolbox.
\newblock {\em Optics express}, 24(22):25129--25147, 2016.

\bibitem{vanaarle2015}
Wim Van~Aarle, Willem~Jan Palenstijn, Jan De~Beenhouwer, Thomas Altantzis, Sara
  Bals, K~Joost Batenburg, and Jan Sijbers.
\newblock {T}he {ASTRA} {T}oolbox: {A} platform for advanced algorithm
  development in electron tomography.
\newblock {\em Ultramicroscopy}, 157:35--47, 2015.

\bibitem{vogel2013}
Jakob Vogel, Tobias Lasser, Jos{\'e} Gardiazabal, and Nassir Navab.
\newblock {T}rajectory optimization for intra-operative nuclear tomographic
  imaging.
\newblock {\em Medical image analysis}, 17(7):723--731, 2013.

\bibitem{weiss2013}
AJ~Weiss, A~Elixhauser, and C~Steiner.
\newblock {R}eadmissions to {US} hospitals by procedure, 2010.
\newblock {\em HCUP Statistical Brief}, 154, 2013.

\bibitem{wu2020}
P.~Wu, N.~Sheth, A.~Sisniega, A.~Uneri, R.~Han, R.~Vijayan, Prasad Vagdargi,
  B.~Kreher, H.~Kunze, G.~Kleinszig, S.~Vogt, S.-F. Lo, N.~Theodore, and J.~H.
  Siewerdsen.
\newblock {M}ethod for metal artifact avoidance in {C}-{A}rm cone-beam {CT}.
\newblock In Guang-Hong Chen and Hilde Bosmans, editors, {\em Medical Imaging
  2020: Physics of Medical Imaging}, volume 11312, pages 522 -- 530.
  International Society for Optics and Photonics, SPIE, 2020.

\bibitem{zaech2019}
Jan-Nico Zaech, Cong Gao, Bastian Bier, Russell Taylor, Andreas Maier, Nassir
  Navab, and Mathias Unberath.
\newblock {L}earning to avoid poor images: {T}owards task-aware {C}-arm
  cone-beam {CT} trajectories.
\newblock In {\em International Conference on Medical Image Computing and
  Computer-Assisted Intervention}, pages 11--19. Springer, 2019.

\bibitem{zhang2007}
Yongbin Zhang, Lifei Zhang, X~Ronald Zhu, Andrew~K Lee, Mark Chambers, and Lei
  Dong.
\newblock {R}educing metal artifacts in cone-beam {CT} images by preprocessing
  projection data.
\newblock {\em International Journal of Radiation Oncology Biology Physics},
  67(3):924--932, 2007.

\bibitem{zheng2011}
Ziyi Zheng and Klaus Mueller.
\newblock {I}dentifying {S}ets of {F}avorable {P}rojections for {F}ew-{V}iew
  {L}ow-{D}ose {C}one-{B}eam {CT} {S}canning.
\newblock In {\em 11th International Meeting on Fully Three-Dimensional Image
  Reconstruction in Radiology and Nuclear Medicine}, 2011.

\end{thebibliography}

%\begin{thebibliography}{}
% etc
%\end{thebibliography}

\end{document}